\documentclass[10pt,journal,compsoc]{IEEEtran}
\ifCLASSOPTIONcompsoc
  \usepackage[nocompress]{cite}
\else
  \usepackage{cite}
\fi
\ifCLASSINFOpdf
\else
\fi
\usepackage{amsmath, amssymb, dsfont}
\usepackage{color, colortbl}
\definecolor{myColor}{rgb}{0.95,0.95,0.95}

\usepackage{wrapfig,lipsum,booktabs}
\usepackage{epsfig}
\usepackage{floatrow}

\usepackage{array}
\newcolumntype{C}[1]{>{\centering\let\newline\\\arraybackslash\hspace{0pt}}m{#1}}
\newcolumntype{L}[1]{>{\let\newline\\\arraybackslash\hspace{0pt}}m{#1}}
\usepackage{multirow}
\usepackage[tableposition=top]{caption}
\usepackage{array,arydshln}
\usepackage{url}
\usepackage{dsfont}
\usepackage{amsfonts}
\usepackage{booktabs}
\usepackage{siunitx}
\usepackage{flushend}

\usepackage[pagebackref=false,breaklinks=true,letterpaper=true,colorlinks,bookmarks=false]{hyperref}

\DeclareMathOperator*{\argmin}{argmin}

\newcommand{\removed}[1]{\textcolor{red}{}}

\newcommand{\oldtext}[1]{} 

\DeclareMathAlphabet{\mathpzc}{T1}{pzc}{m}{n}

\begin{document}
%
\title{Learning Semantic Part-Based Models \\from Google Images}
%
%
%
%

\author{Davide Modolo and Vittorio Ferrari
\IEEEcompsocitemizethanks{\IEEEcompsocthanksitem D. Modolo and V. Ferrari are with the IPAB institute at the University of Edinburgh. E-mail for correspondence: davide.modolo@gmail.com}
}

\IEEEtitleabstractindextext{%
\vspace{-2mm}
\begin{abstract}
We propose a technique to train semantic part-based models of object classes from Google Images.
Our models encompass the appearance of parts and their spatial arrangement on the object, specific to each viewpoint.
We learn these rich models by collecting training instances for both parts and objects, and automatically connecting the two levels. Our framework works incrementally, by learning from easy examples first, and then gradually adapting to harder ones. A key benefit of this approach is that it requires no manual part location annotations.
We evaluate our models on the challenging PASCAL-Part dataset~\cite{chen14cvpr} and show how their performance increases at every step of the learning, with the final models more than doubling the performance of directly training from images retrieved by querying for part names (from 12.9 to 27.2 AP).
Moreover, we show that our part models can help object detection performance by enriching the R-CNN detector with parts.
\end{abstract}
\vspace{-3mm}
 \begin{IEEEkeywords}
 Part Detection, Web Learning, Curriculum Learning \vspace{-1mm}
 \end{IEEEkeywords}
}

\maketitle

\IEEEdisplaynontitleabstractindextext

%
\IEEEpeerreviewmaketitle


%
%
%
%

 

\IEEEraisesectionheading{\section{Introduction}\label{sec:introduction}}
\vspace{-2mm}
\IEEEPARstart{P}{art}-based models have gained significant attention in the last few years. The key advantages of exploiting part representations is that parts have lower intra-class variability than whole objects, they deal better with pose variation and their configuration provides useful information about the aspect of the object. 
Parts localization has therefore been addressed in the context of several vision tasks, such as
object recognition~\cite{chen14cvpr,endres13cvpr,felzenszwalb10pami},
object segmentation~\cite{arbelaez12cvpr,wang15cvpr},
fine-grained classification~\cite{liu12eccv,parkhi12cvpr,zhang14eccv},
human pose-estimation~\cite{liu14eccv,sun11iccv_art,ukita12cvpr},
attribute prediction~\cite{bourdev09iccv,zhang13iccv}, 
action classification~\cite{gkioxari15iccv} and
scene classification~\cite{juneja13cvpr}, achieving state-of-the-art results in many of them.

Part-based methods can be grouped into two sets.
The first set of works define an object part as any patch that is discriminative for the object class~\cite{endres13cvpr,felzenszwalb10pami,arbelaez12cvpr,wang15cvpr,juneja13cvpr}. These works typically discover parts in the training images automatically, without human supervision. However, their resulting parts do not have a meaning for humans (e.g. a patch straddling between the wheel and the chassis of a car~\cite{felzenszwalb10pami}).
The second set of works define parts semantically (e.g. `wheel')~\cite{chen14cvpr,wang15cvpr,liu12eccv,parkhi12cvpr,liu14eccv,sun11iccv_art,ukita12cvpr,zhang13iccv}.
These are more interpretable for a human and are necessary to obtain fine descriptions of objects and their interactions. For example, ``the headlights of the bus are turned on'' and ``the cat is touching the TV with its tail''.
Moreover, part localization is necessary for a robot to correctly grasp an object (e.g. grasp a mug by the handle).
However, existing works on semantic part detection require part location annotations in the training images, which are very expensive to obtain.

In this paper we try to get the best of both worlds by proposing a novel method to train semantic part models of object classes without manual location annotations.
We train these models on images automatically collected from Google Images.
We represent an object class as a mixture over multiple viewpoints. We learn a collection of semantic part
appearance models,
and models of their spatial arrangement on the object,
specific to each viewpoint. 
Moreover, we also train models capable of predicting the viewpoint of the object, which we then use to select an appropriate location model to guide part localization on novel instances. 
%
Learning from the web has been addressed before~\cite{fergus05iccv,vijayanarasimhan:cvpr08,li10ijcv,schroff11pami,liq13cvpr,chen13iccv,divvala14lcvpr,chen15cvpr,novotny16eccvworkshop}, but mostly at the level of object classes. Instead, here we learn complex semantic part-based models from the web.

\begin{figure}[t]
  \centering
    \includegraphics[width=\textwidth]{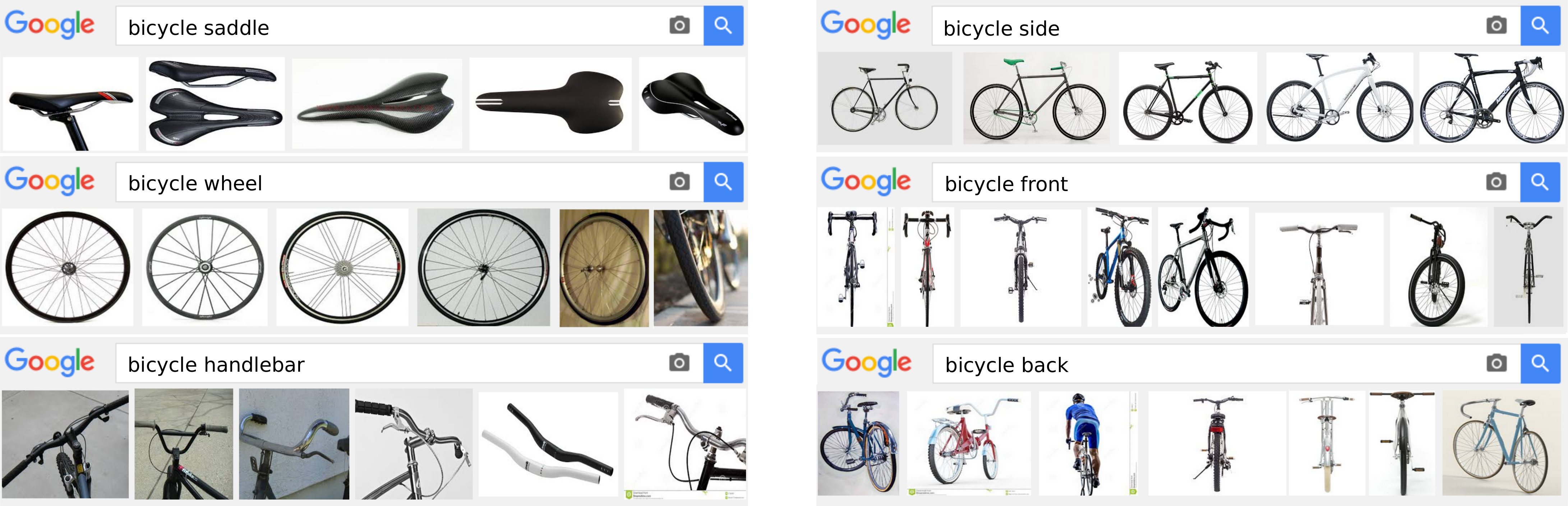}
      \caption{\it \small Images returned from Google Images. On the left examples of queries for object parts, while on the right queries for an object under different viewpoints. Note how the instances are correct, clean and they mostly appear under a uniform background. \vspace{-6mm}} \vspace{-2mm}
      \label{fig:front}
\end{figure}

We learn these rich models fully automatically, entirely from Google Images, by collecting training instances for both \emph{parts} and \emph{objects} (fig.~\ref{fig:front}), and automatically connecting the two levels. 
Our technique incrementally learns from easy examples first, and then gradually adapts to harder examples. This adaptation is done within Google Images, where part images offer easy examples (fig.~\ref{fig:front} left), and then harder examples are mined from object instances (fig.~\ref{fig:front} right). The move from part images to object images also enables us to learn the spatial arrangement of parts on the object (location models).
In a final step, we further adapt our models on an external, non-Google image domain to adapt to even harder examples, e.g. on PASCAL VOC~\cite{everingham10ijcv}.

We demonstrate the effectiveness of our incremental learning algorithm on the PASCAL-Part dataset~\cite{chen14cvpr}.
Interestingly, the performance of our part models increases at every step of the learning, with the final models more than doubling the initial performance (from 12.9 to 27.2 AP).
Moreover, we compare to two other webly-supervised works (LEVAN~\cite{divvala14lcvpr} and NEIL~\cite{chen13iccv}) and show that our part models perform better.
Finally, we also show that our part models can help object detection performance by enriching the R-CNN detector~\cite{girshick14cvpr} with parts. 


\section{Related work}
\label{sec:rel_work}
\vspace{-2mm}
Many works learn non-semantic part models, where the parts are arbitrary patches that are discriminative for an object class~\cite{endres13cvpr,felzenszwalb10pami,arbelaez12cvpr,wang15cvpr,juneja13cvpr}. Our work is more related to semantic part-based models, and to techniques for learning object classes from image search engines.\\

\vspace{-3mm}
\noindent {\bf Semantic part-based models.}
There is a considerable amount of work on using semantic parts to help recognition tasks. 
The largest part of this has focused on the fine-grained recognition problem in several animal domains, such as birds~\cite{liu14eccv,zhang13iccv,zhang14eccv,lin15cvpr} and pets~\cite{liu12eccv,parkhi12cvpr}. In these works, an object is treated as a collection of parts that models its shape and appearance. Semantic parts help capturing subtle object appearance differences that could not be captured by a monolithic object model. These differences are crucial to discriminate between animal breeds. 
Other applications where semantic part-based models have been used are object detection~\cite{chen14cvpr}, articulated human and animal pose estimation~\cite{liu14eccv,sun11iccv_art,ukita12cvpr} and attribute prediction~\cite{zhang13iccv,gkioxari15iccv}. In object detection, parts help dealing with deformed, occluded and low resolution objects. In articulated pose estimation, parts help identifying objects in special configuration (e.g. jumping and sitting) as opposed to canonical ones. Finally, in  attribute prediction, attributes are predicted best by the part containing direct evidence about them.

All the above mentioned methods require accurate part location annotations for training 
(either in terms of keypoints~\cite{liu12eccv,liu14eccv} or bounding boxes~\cite{chen14cvpr,parkhi12cvpr,sun11iccv_art,ukita12cvpr,zhang13iccv,zhang14eccv,lin15cvpr,gkioxari15iccv}). Our framework instead does not require annotations of part positions nor extent, and automatically learns from Google Images instead.\\

\vspace{-3mm}
\noindent {\bf Learning from image search engines.} 
Several works have tried to learn visual models from training samples collected automatically from image search engines~\cite{fergus05iccv,vijayanarasimhan:cvpr08,li10ijcv,schroff11pami,liq13cvpr,chen13iccv,divvala14lcvpr,chen15cvpr,novotny16eccvworkshop}.
Most of them tackle image classification~\cite{fergus05iccv,vijayanarasimhan:cvpr08,li10ijcv,schroff11pami,liq13cvpr} and develop algorithms to find good training samples and learn iteratively.

Some works try to learn object class detectors from the web~\cite{chen13iccv,divvala14lcvpr,chen15cvpr}.
Chen \emph{et al.}~\cite{chen15cvpr} considers only objects and not parts.
LEVAN~\cite{divvala14lcvpr} leverage Google Books Ngrams to discover all appearance variations of an object class, then trains an object detector with a separate component per variation. While some of the components happen to represent parts (e.g. `horse head'), these are treated just like other independent components (at the same level as `jumping horse' and `racing horse').
%
NEIL~\cite{chen13iccv} mines web images to discover common sense relationships between object classes (e.g. `car is found in raceway'), including also some part-of relations (e.g. `wheel is part of car').
%
%
Importantly, both LEVAN and NEIL learn simple object class detectors, consisting of `root filters' only. Instead we learn more complex, structured models of object classes, which include semantic part appearance models and their spatial arrangements within the object, conditioned on object viewpoint.
Note how~\cite{chen13iccv,divvala14lcvpr,chen15cvpr} do not report quantitative localization results for part detection.
Finally, we believe our work is complementary to~\cite{chen13iccv,divvala14lcvpr,chen15cvpr}. The frameworks of~\cite{chen13iccv,divvala14lcvpr} could provide a list of which parts belong to which object class, which could be passed on to our technique to learn more complex models.
Moreover, our part models could be used in combination with the strong R-CNN root filters learned from the web by~\cite{chen15cvpr} (analog to sec.~\ref{sec:det}). 

The concurrent work \cite{novotny16eccvworkshop} is the only other one to learn semantic part models from the web. It learns the structure of objects in an embedding space where geometric relationships are implicitly conveyed by non-semantic mid-level parts. Instead, we learn explicit relationships on the semantic parts themselves, based on object and parts instances automatically mined from web images.

\begin{figure*}[t]
  \centering
    \includegraphics[width=\textwidth]{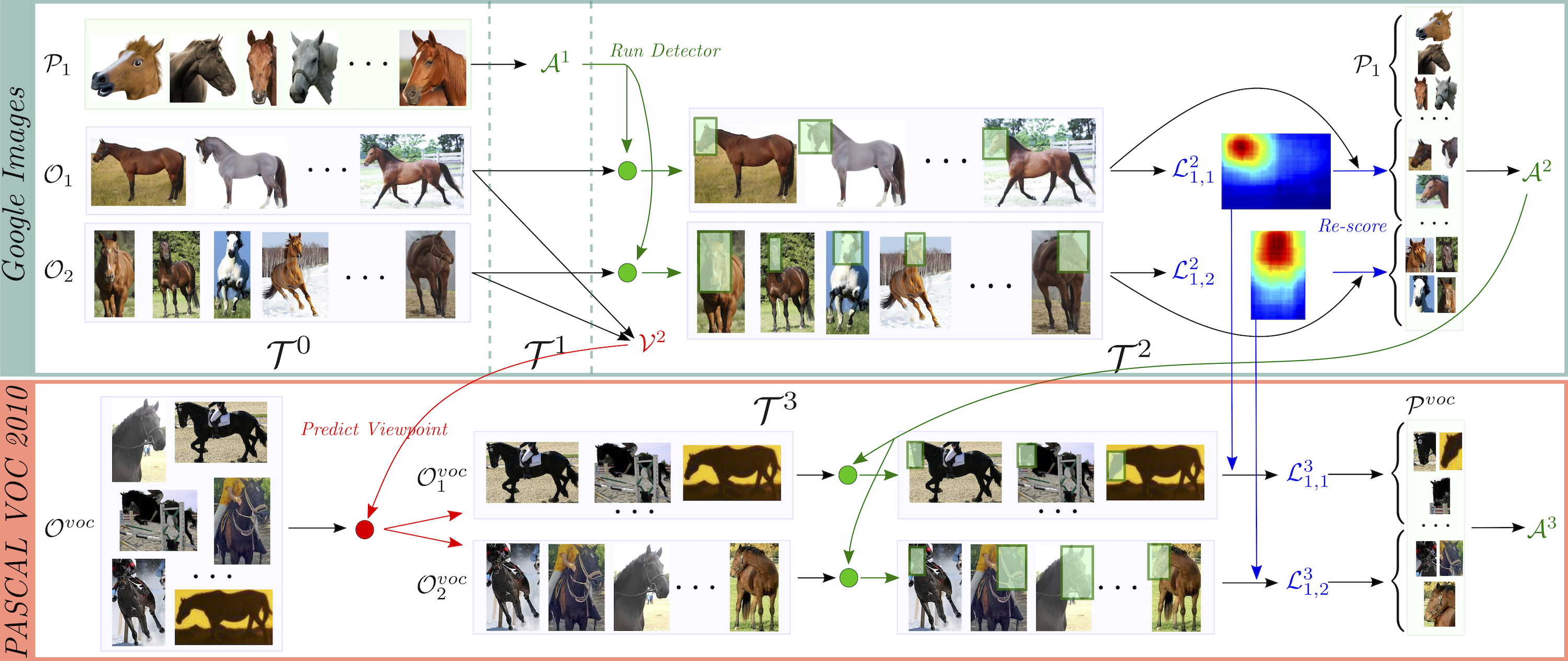}
      \caption{\it \small Schema of our framework for object class `horse', simplified to just one part `head' and two viewpoints `left' and 'front'. At ${\mathcal{T}^0}$ the framework downloads horse instances (${\mathcal{O}_1}$ and ${\mathcal{O}_2}$, for the two viewpoints) and head instances (${\mathcal{P}_1}$) from Google Images. At ${\mathcal{T}^1}$ it learns a first head appearance model ${\mathcal{A}^1}$. At ${\mathcal{T}^2}$ it then hunts for new head instances from the horse images in ${\mathcal{O}_1}$ and ${\mathcal{O}_2}$ to first train two head location models ${\mathcal{L}^2_{1,1}}$ and ${\mathcal{L}^2_{1,2}}$, one for each horse viewpoint, and later to re-train a more accurate head appearance model ${\mathcal{A}^2}$. Finally, it also learns a viewpoint classifier ${\mathcal{V}^2}$. At ${\mathcal{T}^3}$ it then predicts the viewpoint of objects in ${\mathcal{O}_{voc}}$ using ${\mathcal{V}^2}$ and hunts for more part instances from them. These are then used to train our final part appearance model ${\mathcal{A}^3}$ and part location models  ${\mathcal{L}^3_{1,1}}$ and ${\mathcal{L}^3_{1,2}}$. Note how at time ${\mathcal{T}^1}$ the framework has only seen part instances and has no information to learn neither $\mathcal{L}^1$ nor ${\mathcal{V}^1}$. \vspace{-4mm}} \vspace{-3mm}
      \label{fig:schema}
\end{figure*} 

\vspace{-2mm}
\section{Overview of our approach} \label{sec:overview}
\vspace{-1mm}
We present here an overview of our framework for automatically learning compositional semantic part models. We learn these models for each object class separately.
For each class, we use Google Images to collect images of the object under several pre-defined viewpoints, and images of its parts. 
We use the part samples to train initial part appearance models, which are later used to learn the connection between the parts and the whole object. Learning this association is the key to our compositional part models. 

For each class, we learn part appearance models $\mathcal{A}$, part location models $\mathcal{L}$, and object viewpoint classifiers $\mathcal{V}$. 
Our framework operates in four stages: $\mathcal{T}_0 - \mathcal{T}_3$ (fig.~\ref{fig:schema}). In the first stage $\mathcal{T}_0$, we collect training samples from Google Images (objects and parts, fig.~\ref{fig:front}). Then, we iteratively learn the components of our part models, each time learning from harder examples:
($\mathcal{T}_1$) images containing only parts from Google,
($\mathcal{T}_2$) part examples mined from object images from Google, and
($\mathcal{T}_3$) part examples mined from object images from the PASCAL VOC 2010 dataset~\cite{everingham10ijcv}.
Every stage is fully automatic and does not require human intervention.

For each object part, we learn one appearance model and $V$ location models, one for each viewpoint in our predefined set. A single location model is not sufficient to capture the position of a part with respect to the object, as this is strongly affected by viewpoint changes. For example, the front view of a bicycle has one wheel on the bottom-center of the bicycle, while a bicycle from the side has two wheels on the bottom left and right (fig.~\ref{fig:front}, right).

For simplicity, in the rest of the paper we use superscripts to indicate the stage a model component is trained at. For example, our part appearance model $\mathcal{A}^2$ is trained at stage $\mathcal{T}^2$, while $\mathcal{A}^3$ at stage $\mathcal{T}^3$. \\

\vspace{-2mm}
\noindent {\bf $\mathbf{\mathcal{T}^0}$: Collecting data.}
Our framework queries Google Images for images of an object under canonical viewpoints and images of each of its parts (sec.~\ref{sec:data}). 
%
These images are biased towards simple representations, in a uniform background and they reliably contain the wanted object (or part). However, one image may contain multiple instances or objects not appearing nicely in the centre (fig~\ref{fig:segm}).
It would be better if each object/part instance would be enclosed in a tight bounding-box.
Bounding-boxes around parts help learning accurate appearance models as they exclude background pixels, and around objects they help learning accurate part location models as they provide a stable coordinate frame common to all instances.
We therefore devise a simple, yet effective algorithm to fit a tight bounding-box around each part/object instance (sec.~\ref{sec:data}). Finally, we consider each bounding-box as a separate image, obtaining our initial training set.
We denote with $\mathcal{O}_j \in \mathcal{O}$ the set of images of the object under viewpoint $v_j$ and with $\mathcal{P}_i \in \mathcal{P}$ the set of images of part $p_i$.\\


\vspace{-3mm}
\noindent {\bf $\mathbf{\mathcal{T}^1}$: Learning from Google's easy examples.}
%
For each part $p_i$, our framework learns an appearance model $\mathcal{A}^1_i$ on the part images $\mathcal{P}_i$ (sec.~\ref{sec:app}). 
These are the easiest samples, as in these images the part often appears isolated from the object and against a clean background (fig.~\ref{fig:front} left).\\
%


%

\vspace{-3mm}
\noindent {\bf $\mathbf{\mathcal{T}^2}$: Learning from Google's harder examples.}
In this stage our framework moves on to object images $\mathcal{O}$. It learns part location models $\mathcal{L}^2$ and updates all part appearance models by using additional samples from $\mathcal{O}$ (sec.~\ref{sec:app}). Moreover, it trains an object viewpoint classifier $\mathcal{V}^2$ on $\mathcal{O}$ (sec.~\ref{sec:view}). 

For each viewpoint $v_j$ and part $p_i$, it learns $\mathcal{L}^2_{i, j}$. The key idea is to run $\mathcal{A}^1_i$ on the object images $\mathcal{O}_j$. The top-scored part detections are likely to be correct and, importantly, they are now {\em localized within an object image}. Therefore, they provide valuable training samples for the location of the part within the object (sec.~\ref{sec:loc}).
%
The intuition here is that objects captured under the same viewpoint have parts in similar spatial arrangements. For example, all horses from the side have the head on the left side of the image, mostly on top (fig.~\ref{fig:schema}).
Since all objects in $\mathcal{O}_j$ are in the same viewpoint $v_j$, correct detections of a part will be consistently found at similar locations across different object instances. 

Subsequently, the framework mines part samples automatically from each $\mathcal{O}_j \in \mathcal{O}$ using $\mathcal{A}^1_i$ and the corresponding $\mathcal{L}^2_{i,j}$ (sec.~\ref{sec:mining}). 
The process looks for detections that have a high score according to $\mathcal{A}^1_i$ and are at the right location according to $\mathcal{L}^2_{i,j}$. By combining these two source of information, we consistently discover correct part samples.
Finally, the framework uses these samples to update part appearance models to $\mathcal{A}^2_i$. Note how these new samples are more difficult than the ones in $\mathcal{T}^1$, since come from images showing whole objects and against natural backgrounds.
Lastly, the framework trains an object viewpoint classifier $\mathcal{V}^2$ on $\mathcal{O}$, by using each set of images $\mathcal{O}_j \in \mathcal{O}$ as training set for viewpoint $v_j$ (sec.~\ref{sec:view}).
Finally, note how at this stage the framework has trained a complete, rich model (part appearance $\mathcal{A}^2$, part location $\mathcal{L}^2$, object viewpoint $\mathcal{V}^2$) entirely and automatically from Google Images (fig.~\ref{fig:schema}, top). \\ \removed{The key is to associate parts to their object and learn the connections.}

%


\vspace{-3mm}
\noindent {\bf $\mathbf{\mathcal{T}^3}$: Learning from PASCAL VOC.}
In this final stage the framework refines all $\mathcal{A}^2$ and $\mathcal{L}^2$ using even more difficult training samples from another domain (sec.~\ref{sec:app},~\ref{sec:loc}). 
These samples are mined automatically from the PASCAL VOC dataset, which contains photographs depicting challenging objects in natural scenes, often occluded or truncated (fig.~\ref{fig:schema}, bottom). These are much harder than the ones in stage $\mathcal{T}^2$, where each image had a single whole object.
The framework mines positives as in step $\mathcal{T}^2$, but using the updated $\mathcal{A}^2$ instead of the initial $\mathcal{A}^1$ (sec.~\ref{sec:mining}).
Similarly to other works~\cite{parkhi12cvpr,lin15cvpr}, we only search for parts inside the ground-truth bounding-boxes of the object class (which are provided with PASCAL VOC). We call this set $\mathcal{O}^{voc}$. Furthermore, we call the set of mined positives $\mathcal{P}^{voc}$. 
In order to use our viewpoint-specific location models, we need to determine the viewpoint of the images in $\mathcal{O}^{voc}$.
We automatically predict $v_j$ for each image in $\mathcal{O}^{voc}$ using the object viewpoint classifier $\mathcal{V}^2$. 
%
%
Finally, after mining new positives, the framework finally trains final location models $\mathcal{L}^3_{i, j}$ and part appearance models $\mathcal{A}^3_i$ (sec.~\ref{sec:mining}).  

\vspace{-1mm}
\section{The components of our approach} \label{sec:tech}

We detail below the components of our approach. In sec.~\ref{sec:data} we describe our data collection mechanism. In sec.~\ref{sec:app},~\ref{sec:loc} and~\ref{sec:view} we describe how to train $\mathcal{A}$, $\mathcal{L}$ and $\mathcal{V}$, respectively. In sec.~\ref{sec:mining} we then present our procedure to automatically mine new part instances from objects.

\vspace{-3mm}
\subsection{Data collection and preprocessing} \label{sec:data}
\vspace{-1mm}
This section describes how we download part and object images from Google and how we fit a tight bounding-box around each part/object instance in them.\\

\vspace{-3mm}
\noindent{\bf Querying Google Images.} 
We collect images of an object under multiple viewpoints and of its parts (fig.~\ref{fig:front}). 
We keep the top 100 retrieved images for each object viewpoint and the top 25 for each object part. We observed these numbers to produce good, clean images. Collecting more than 25 part images sometimes delivers spurious images without the part, which would introduce noise in the learning process.


For each object class, we use the names of its parts as listed in the PASCAL-Part Dataset~\cite{chen14cvpr} and the viewpoint names specified by PASCAL VOC 2010~\cite{everingham10ijcv} (\emph{front, back, left, right}). As {\emph{left}} and {\emph{right}} is not a level of granularity satisfied by Google Images yet, 
we query for a generic {\emph{side}} viewpoint (fig.~\ref{fig:front} right-top) and then automatically split the retrieved images into left and right subsets.
In order to do this, we first augment the image set by mirror flipping all images horizontally, and then we cluster them by minimizing the intra-cluster HOG compactness, similarly to~\cite{felzenszwalb10pami}. 



\begin{figure}[t]
  \centering
    \includegraphics[width=\textwidth]{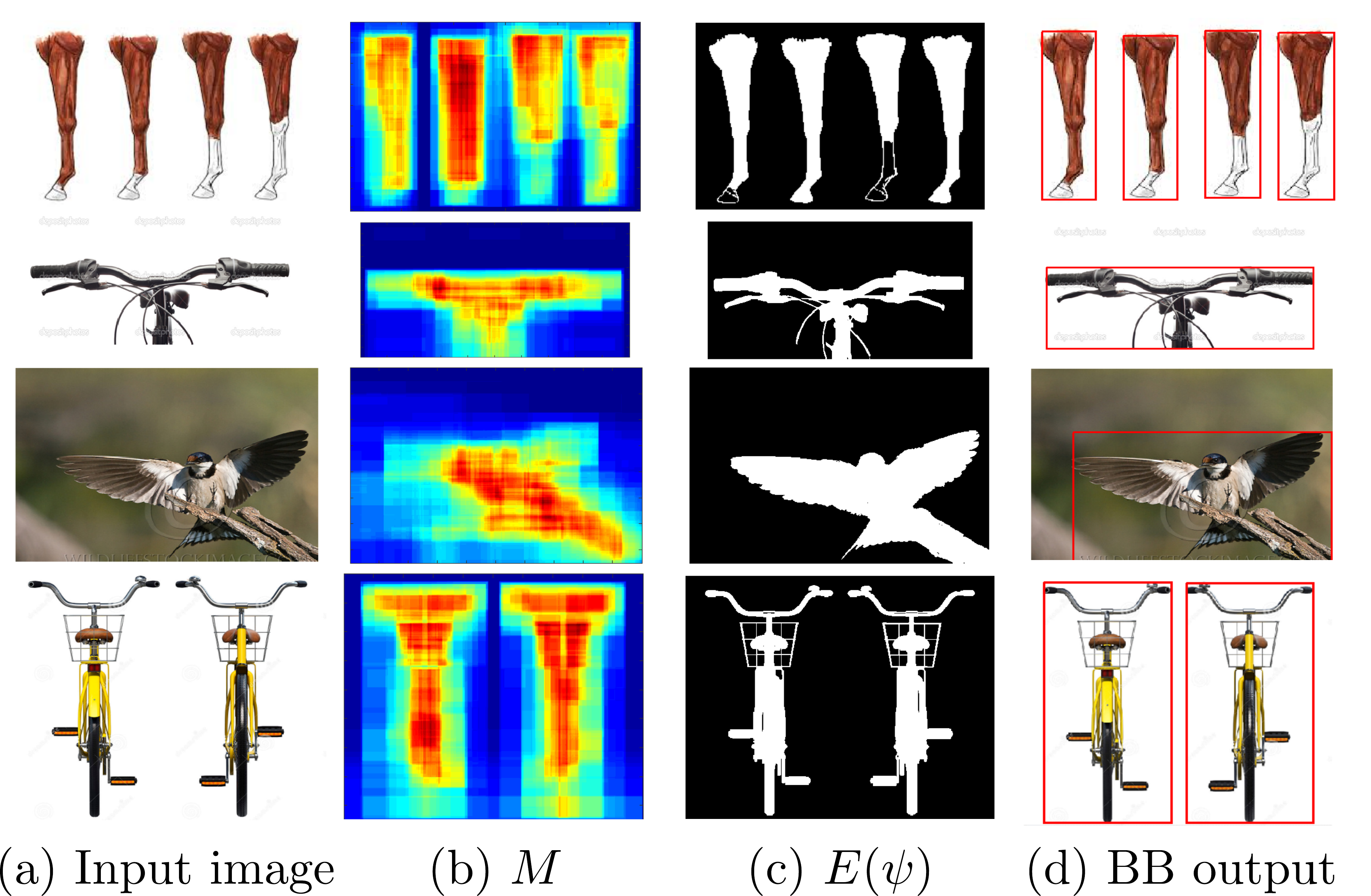}
      \caption{\it \small Examples of the steps of our procedure to fit bounding-boxes to object/part instances in web images (sec.~\ref{sec:data}). (a) is the input image; (b) is the initial rough foreground estimate $M$; (c) is the output of the segmentation process; and (d) are the bounding-boxes fit to connected components in the segmentation.
      Note how the two images `horse leg' and `bicycle front' have multiple part/object instances and our method is able to fit a separate bounding-box around each of them.\vspace{-4mm}} \vspace{-2mm}
      \label{fig:segm}
\end{figure}

\noindent {\bf Fitting bounding-boxes.} 
As mentioned in sec.~\ref{sec:overview}, we want to fit a tight bounding-box around each object/part instance. These bounding-boxes
help learning accurate appearance and location models for the parts.
Fortunately, Google Images results are biased towards whole objects in a uniform background (fig.~\ref{fig:segm}a) and unoccluded. These are easy to localize. 
We formulate this task as a pixel labelling problem, where each pixel $\phi_i$ can take a label $l_i \in \{0, 1\}$ (background or foreground).
We aim at finding the best labelling $\psi^* = \argmin_L E(\mathcal{\psi})$. Similar to other segmentation works \cite{Rother04-tdfixed,kuettel12cvpr,RosenfeldICCV11}, we define an energy function: 
\begin{equation}
\small
\vspace{-1mm}
E(\mathcal{\psi}) = \sum_i M_i(l_i) + \sum_i G_i(l_i) + \alpha \sum_{i, j} V(l_i, l_j),
\label{eq:Seg}
\end{equation}
where, the pairwise potential $V$ encourages smoothness by penalising neighbouring pixels taking different labels and the unary potential $G_i$ evaluates how likely a pixel $i$ is to take label $l_i$ according to an appearance model which consists of two GMMs~\cite{Rother04-tdfixed} (foreground and background). 
Inspired by \cite{kuettel12cvpr}, we produce an initial rough estimate $M$ of which pixels lie on the object, and use it both to estimate the appearance models $G$, and as a unary potential of its own. 
%
%
We do this in an unsupervised manner, based purely on the spatial distribution of object proposals \cite{uijlings13ijcv} in the image (fig.~\ref{fig:segm}b).
We define the likelihood $M_i(1)$ of a pixel $i$ to be foreground as the number of proposals that contain it, divided by the total number of proposals in the image ($M_i(0) = 1-M_i(1)$).
The idea is that if a pixel is contained in many proposals, then it is likely to belong to the object. 

As in GrabCut~\cite{Rother04-tdfixed}, we iteratively alternate between minimizing the energy (eq.~\ref{eq:Seg}) to obtain a segmentation, and updating the appearance models based on this segmentation.
After a few iterations this process converges and we fit a tight bounding-box around each connected component in the final segmentation (fig.~\ref{fig:segm}d).
We apply this procedure to all images collected from Google, obtaining the initial training set of object images $\mathcal{O}$ and part images $\mathcal{P}$ (treating each bounding-box as a separate image).\\

\vspace{-3mm}
\noindent{\bf Part proposals.}
We generate class-independent part proposals inside each image in $\mathcal{O}$ using~\cite{uijlings13ijcv}.
As observed by Zhang \emph{et al.}~\cite{zhang14eccv}, these proposals achieve low recall on small semantic parts. In order to overcome this difficulty, we changed the standard settings of~\cite{uijlings13ijcv} to return smaller proposals and increase part recall. 
This results in about 2000 proposals per object image in $\mathcal{O}$, likely to cover all parts.
In the rest of the paper we use $\mathcal{W}$ to refer to the set all part proposals over all object images.

\subsection{Training part appearance models $\mathcal{A}$}
\label{sec:app}
\vspace{-1mm}
Each stage of our learning framework updates the part appearance models of the object class. We describe here how these models are trained at each stage.

\noindent{\bf Stage $\mathcal{T}^1$.}
We train $\mathcal{A}^1$ on the image set $\mathcal{P}$, containing simple part images. As appearance model we use a convolutional neural network (CNN) and train it to distinguish between the $P$ parts.
More specifically, we start from AlexNet pre-trained on the ImageNet 2012 classification challenge~\cite{krizhevsky12nips} and replace its original 1000-way $\mathtt{fc8}$ classification layer with a $P$-way $\mathtt{fc8}$ layer.
We then finetune the whole network for part classification on the images in $\mathcal{P}$.
Note how $\mathcal{P}$ only contains 25 samples per part. In order to avoid overfit we use a learning rate of $10^{-4}$ and apply early stopping ($1000$ iterations, 5 epochs). Higher learning rates cause the parameters to vary abruptly over iterations, whereas $10^{-4}$ results in a smooth learning curve.
At test time we use the softmax at layer $\mathtt{fc8}$ to predict how likely a proposal is to contain each of the parts.
At all times we use the publicly available CNN implementation~\cite{jia13caffe}. 

\noindent{\bf Stage $\mathcal{T}^2$.}
At this stage we learn $\mathcal{A}^2$ on a larger training set of examples from both part images and part samples automatically mined from object images using the appearance model from stage $\mathcal{T}^1$ and the location model from stage $\mathcal{T}^2$ (sec.~\ref{sec:mining}, fig.~\ref{fig:schema}).
As appearance model we train a similar CNN to the one of stage $\mathcal{T}^1$, but with a difference: we use a richer $(P+1)$-way $\mathtt{fc8}$ layer, where the additional output is used to classify background patches. Note how by mining for positive part instances in object images (sec.~\ref{sec:mining}) we indirectly discover negative proposals (those with intersection-over-union ($IoU$)~\cite{everingham10ijcv} $\le 0.3$ with mined positives).


\noindent{\bf Stage $\mathcal{T}^3$.}
In the last stage we train $\mathcal{A}^3$ on the harder image set $\mathcal{P}^{voc}$, using as training samples parts automatically mined from $\mathcal{O}^{voc}$ using the part detector from stage $\mathcal{T}^2$ (sec.~\ref{sec:mining}).
As appearance model we train a CNN as in $\mathcal{T}^2$.

\begin{figure}[t]
  \centering
    \includegraphics[width=1\textwidth]{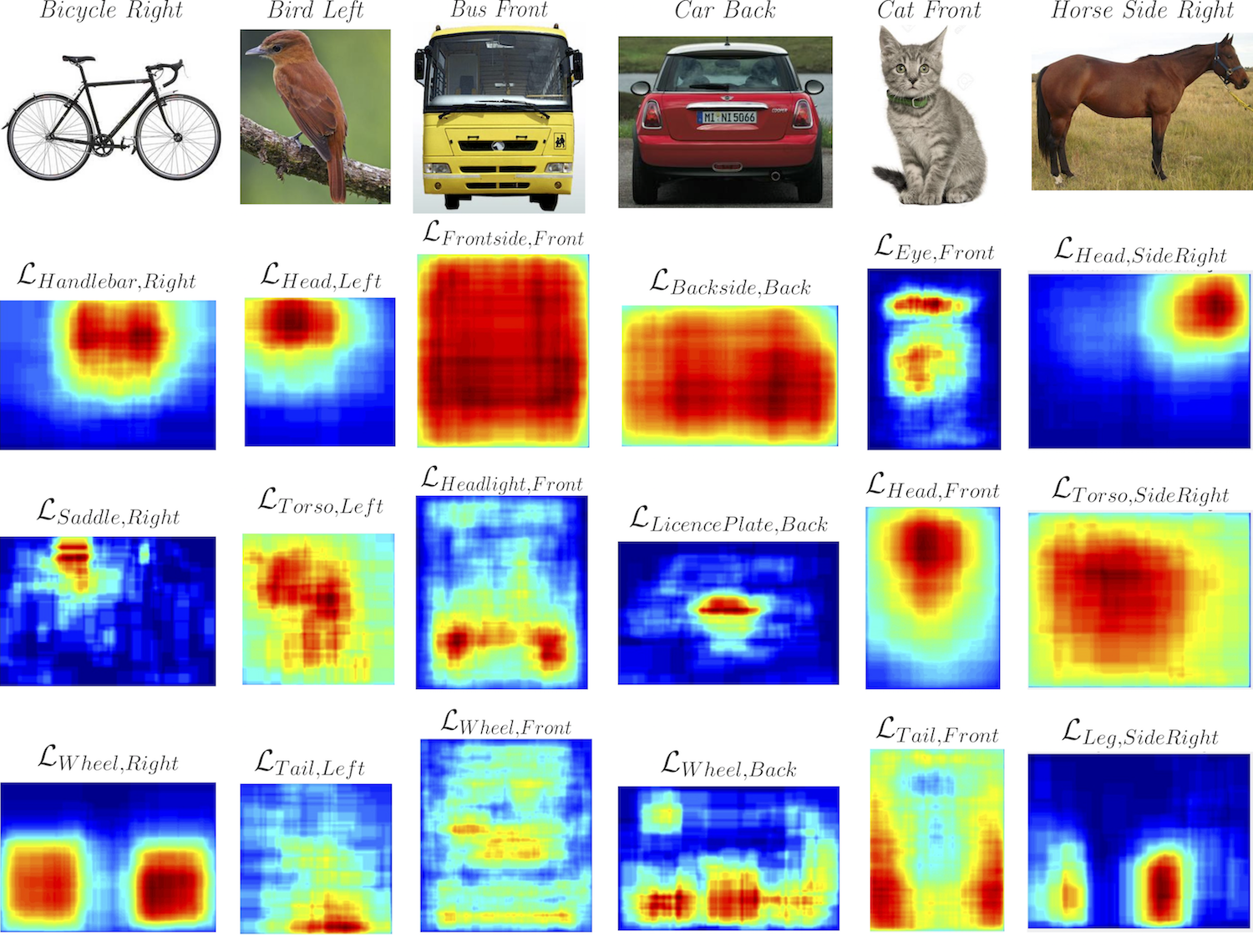}
      \caption{\it \small Examples of our location models $\mathcal{L}$. We show a canonical image of each object captured under one of our viewpoints and the location models of their parts. These models nicely capture the average position of each part within the object in that viewpoint. Note how these are automatically learnt from Google Images.
      For visualization, we show a 2D projection of the location models, which however live in a 4D space defined not only by the $(x,y)$ position of a proposal, but also by its scale and aspect ratio.\vspace{-5mm}} \vspace{-4mm}
      \label{fig:loc}
\end{figure}

\vspace{-4mm}
\subsection{Learning part location models $\mathcal{L}$} \label{sec:loc}
The appearance model $\mathcal{A}$ scores part proposals in an image based on their appearance only.
We build location models to capture complementary knowledge about likely positions and scales of the object parts within the the coordinate frame of the object.
In stage $\mathcal{T}^2$ we learn the location models purely from Google Images and in stage $\mathcal{T}^3$ we adapt them to a different domain.
In this subsection we use $\mathcal{W}_j$ to refer to the set of all part proposals in object images $\mathcal{O}_j$. \removed{, i.e. under viewpoint $v_j$.}\\

\vspace{-3mm}
\noindent{\bf Part training samples.}
For each viewpoint $v_j$ and part $p_i$, we learn a separate location model $\mathcal{L}_{i,j}$ from a set of training proposals $\mathcal{W}_{j,i} \in \mathcal{W}_j$ likely to contain part $p_i$.
We describe here how we acquire these part samples $\mathcal{W}_{i, j}$ at stage $\mathcal{T}^2$, i.e. from object images $\mathcal{O}_j$. The key idea is to run the part detector $\mathcal{A}^1_i$ on these images and retain the top-scored part detections. As these detections are localized within an object image, they provide examples of the location of the part within the object.
More precisely, for each image we score all part proposals with the appearance model $\mathcal{A}^1_i$, perform non-maximum suppression, and pick up to 3 detections per image (the top scored ones, if they score above a minimum confidence threshold). These detections form $\mathcal{W}_{i, j}$.
This way of picking detections strikes a good trade-off between keeping all correct locations, but without including too many false-positives.
At $\mathcal{T}^3$, we enrich the sample set $\mathcal{W}_{j, i}$ with the top detections produced by running the appearance model $\mathcal{A}^3_i$ on $\mathcal{O}^{voc}$. More specifically, each of these detections gets assigned to the $\mathcal{W}_{j, i}$ of viewpoint predicted by $\mathcal{V}^2$.
These new samples are used to train the refined location models $\mathcal{L}^2_{i,j}$.\\

\vspace{-3mm}
\noindent{\bf Training a location model.}
The location model $\mathcal{L}_{i,j}$ scores on an input part proposal $w'$ by the density of the training set $\mathcal{W}_{j,i}$ at $w'$
\begin{equation}
\small
\vspace{-1mm}
\mathcal{L}_{i,j}(w') = \frac{1}{|\mathcal{W}_{j,i}| \cdot h}\sum_{w \in \mathcal{W}_{j,i}} K \left(\frac{D(w',w)}{h}\right)
\label{eq:lm}
\end{equation}
where $D(w',w)$ is distance between two proposals: $D(w',w) = 1 - \frac{w' \cap w}{w' \cup w}$ and $K(u)$ is the uniform density function $K(u) = \frac{1}{2} \mathds{1}(|u| \le 1)$.
\removed{\begin{equation}
\small
\vspace{-1mm}
K(u) = \frac{1}{2} \mathds{1}(|u| \le 1)
\end{equation}}

In this formulation $\mathcal{L}_{i,j}(w')$ has an intuitive interpretation as the percentage of proposals in $\mathcal{W}_{j,i}$ which are close to $w'$ ($IoU < h$).
If many training proposals are near $w'$, then $\mathcal{L}_{i,j}(w')$ is large, indicating that $w'$ is likely to contain the part. Conversely, if only a few proposal are near $w'$, then $\mathcal{L}_{i,j}(w')$ is small, indicating it is more likely to cover a background patch.
The bandwidth $h$ controls the degree of smoothing and in our experiments we set it to $h=0.5$.

Note how $D(w',w)$ compares part proposals across different images. For this to be meaningful, $D$ operates in a coordinate frame common to all images in $\mathcal{O}_j$ (by normalizing it by the average width and height of all images). This normalization is specific to a viewpoint $v_j$, so it preserves its aspect-ratio.\\

\vspace{-3mm}
\noindent{\bf Model behaviour.}
Fig.~\ref{fig:loc} shows examples of some location models learned at stage $\mathcal{T}^2$.
Thanks to the way we build them, our location models are robust to errors in the training set: correct training samples tend to cluster around the right locations of a part, whereas incorrect ones tend to scatter across the whole object. This results in strong peaks at the correct locations in the model, with only lower values everywhere else (e.g. headlight for Bus Front).
Moreover, note how our location model is suitable for a variety of cases.
Unique parts of rigid objects form unimodal distributions (Bicycle saddle, Car license plate),
while Bicycle wheel and Horse leg form bimodal ones.
Even in the hard case of highly movable parts of deformable object classes (e.g. Cat tail), the model learns that they can appear over broader regions and spreads the density accordingly.

\vspace{-3mm}
\subsection{Training the viewpoint classifier $\mathcal{V}^2$}
\label{sec:view}
During stage $\mathcal{T}^2$ we train classifiers $\mathcal{V}^2$ on $\mathcal{O}$ for predicting the viewpoint of the object in an image.
We train a CNN to distinguish between the four viewpoints (\emph{front, back, left, right}) for which we collected object images from Google in sec.~\ref{sec:data} (fig.~\ref{fig:front} right). We used these images to train $\mathcal{V}^2$ and,
as for $\mathcal{A}$ (sec.~\ref{sec:app}), we took the CNN pre-trained on the ImageNet classification challenge and replaced its original 1000-way $\mathtt{fc8}$ classification layer with a 4-way $\mathtt{fc8}$.
During stage $\mathcal{T}^3$, the viewpoint classifier is useful to select an appropriate location model for object images $\mathcal{O}^{voc}$.
Given an input image, we select the viewpoint with the highest probability.
Note that the PASCAL VOC 2010 dataset has manual viewpoint annotations for some objects ($\sim60\%$ for the classes we consider). We use these annotations in sec.~\ref{sec:res_view} to evaluate how well our viewpoint classifier $\mathcal{V}^2$ works. 

\vspace{-2mm}
\subsection{Mining for new part instances}
\label{sec:mining}
In stages $\mathcal{T}^2$ and  $\mathcal{T}^3$ we mine for new part instances in $\mathcal{O}$ and $\mathcal{O}^{voc}$, respectively. For simplicity, we describe the process to mine from  $\mathcal{O}$. 
Given each set of images $\mathcal{O}_j \in \mathcal{O}$ showing viewpoint $v_j$, we mine positives for part $p_i$ using the appearance model $\mathcal{A}^1_i$ and the location model $\mathcal{L}^2_{i, j}$. For each image, we score all its part proposals with $\mathcal{A}^1_i$ and $\mathcal{L}^2_{i, j}$, perform non-maximum suppression and keep only the proposals with high score. We repeat this for all $\mathcal{O}_j \in \mathcal{O}$ and obtain our final set of new samples.
Importantly, we mine for new part instances within object bounding boxes only. Even though the initial appearance models $\mathcal{A}^1$ were trained on 25 samples only, they still manage to localize new part instances, as the search space is very limited.
Moreover, this process is able to mine new instances that look significantly different than those in the initial set of easy examples $\mathcal{P}$ (e.g. a frontoparallel wheel against a white background vs a out-of-plane rotated wheel on an actual car, fig.~\ref{fig:res}), as new part instances can be selected if at the right location according to $\mathcal{L}$ , even when $\mathcal{A}$ is not confident about them.

Mining from $\mathcal{O}^{voc}$ is analogous, but requires an extra step, where we use the viewpoint classifier $\mathcal{V}^2$ (sec.~\ref{sec:view}) to predict the otherwise unknown viewpoints of objects $\mathcal{O}^{voc}$.

\begin{table}
\caption{\small Viewpoint classification results (average precision). \vspace{-3mm}}
\label{table:view}
\footnotesize
\resizebox{1\columnwidth}{!}{
\begin{tabular}{c : c c c c c c c c c : c}
\toprule
& {\it Bicycle} & {\it Bird} & {\it Bus} & {\it Car} & {\it Cat} &{\it Cow} & {\it Dog} & {\it Horse} & {\it Sheep} & mean \\\midrule
$\mathcal{V}^2$ & 51.0 & 48.1 & 58.4 & 43.2 &  59.0 & 61.1 & 52.4 & 53.3 & 57.3 & \bf 53.7 \\
 $\mathcal{V}^{FS}$ & 42.6 & 39.3 & 57.8 & 38.6 & 55.4 & 57.9 & 51.8 & 44.5 & 55.9 & 49.3\\
\bottomrule
\end{tabular}}
\vspace{-3mm}
\end{table}

\vspace{-3mm}
\section{Experiments and conclusions}
\label{sec:exp}
\vspace{1mm}
\subsection{Datasets}
\vspace{-1mm}
We evaluate our framework and all its intermediate stages on the recent PASCAL-Part dataset~\cite{chen14cvpr}, which augments PASCAL VOC 2010~\cite{everingham10ijcv} with pixelwise semantic part annotations.
For evaluation we fit a bounding-box to each part segmentation mask. 
Finally, the dataset contains a \texttt{\small train} and a \texttt{\small validation} subsets. 
We mine new part instances from \texttt{\small train} in stage $\mathcal{T}^3$, and measure the performance of our framework on \texttt{\small validation}. We verified by using a near-duplicate detector that none of the images we collected from Google Images are in Pascal Parts. 

We evaluate on nine diverse object classes ({\em bicycle, bird, bus, car, cat, cow, dog, horse, sheep}), three parts each (table~\ref{table:res}).
We treat each leg as a separate instance, rather than grouping them into a `super-part' (as done by~\cite{chen14cvpr}).
Note how previous works evaluating on PASCAL-Parts consider fewer classes/parts~\cite{chen14cvpr,wang15cvpr,hariharan15cvpr} and operate in a fully supervised scenario (training from manual part location annotations).

\vspace{-1mm}
\subsection{Viewpoint prediction}
\label{sec:res_view}
\vspace{-1mm}
We evaluate here our viewpoint classifier $\mathcal{V}^2$ trained purely from Google Images (sec.~\ref{sec:view}). We compare it against a viewpoint classifier $\mathcal{V}^{FS}$ trained using manual annotations from PASCAL VOC 2010 \texttt{\small train}. In both cases we use the same CNN model and training procedure (sec.~\ref{sec:view}).
We evaluate both classifiers in terms of accuracy on \texttt{\small validation} (table~\ref{table:view}). 
Results show that our viewpoint classifier $\mathcal{V}^2$ considerably outperforms the fully supervised classifier $\mathcal{V}^{FS}$. Results are not surprising, as objects in the PASCAL VOC dataset appear often truncated or occluded and sometimes labelled with the wrong viewpoint. Instead, the images from Google have clean objects with well defined viewpoints (fig.~\ref{fig:front}). Moreover, objects appear as a whole, leading to better prediction performance.

\begin{table*}
\caption{\small Part detection results (average precision) on the \texttt{\small validation} set of PASCAL-Part dataset. \vspace{-2mm}} \vspace{-2mm}
 \label{table:res}
\centering
{
\tiny
\resizebox{\columnwidth}{!}{
\begin{tabular}{| l l | c : c : ccc : ccc | cc:cc|} 
\hline
\multicolumn{1}{|c}{} & & $\mathcal{T}^0$ & $\mathcal{T}^1$ & \multicolumn{3}{:c}{$\mathcal{T}^2$} &  \multicolumn{3}{:c|}{$\mathcal{T}^3$} & \multirow{2}{*}{$\mathcal{A}^{FS1}$} & \multirow{2}{*}{$\mathcal{A}^{FS2}$} &  LEVAN  & NEIL \\
& & $\mathcal{A}^0$ & $\mathcal{A}^1$ & $\mathcal{A}^1+\mathcal{L}^2$ & $\mathcal{A}^2$ & $\mathcal{A}^2+ \mathcal{L}^2$ & $\mathcal{A}^3$ & $\mathcal{A}^3+ \mathcal{L}^2$ & $\mathcal{A}^3+ \mathcal{L}^3$ & & & \cite{divvala14lcvpr} & \cite{chen13iccv} \\ 
\hline
\multirow{3}{*}{\it Bicycle} & \it Wheel & 37.2 & 39.6 & 50.5 & 53.9 & 58.7 & 56.6 & 64.0 & 63.9& 75.7 & 74.2 & 24.7 & 43.1\\ 
& \it Saddle & 4.2 & 9.8 & 14.3 & 14.0 & 14.5 & 17.2 & 20.6 & 20.7 & 35.5 & 31.7 & - & -\\ 
& \it Handlebar & 2.2 & 5.9 & 3.5 & 5.9 & 5.9 & 8.5 & 8.7 & 9.9 & 25.6 & 21.1 & - & -\\ 
\hline
\multirow{3}{*}{\it Bird} 
& \it Head & 16.4 & 16.4 & 13.5 & 22.4 & 21.0 & 22.7 & 22.6 & 22.7 & 55.3 & 52.0 & - & -\\
& \it Torso & 2.6 & 5.2 & 10.1 & 38.0 & 48.4 & 48.9 & 53.7 & 55.5 & 60.8 & 56.3 & - & -\\ 
& \it Tail & 0.2 & 0.8 & 2.0 & 0.5 & 0.5 & 1.4 & 2.1 & 2.0 & 8.7 & 5.1 & - & -\\ 
\hline
\multirow{3}{*}{\it Bus} 
& \it Frontside & 40.6 & 43.1 & 59.5 & 60.5 & 68.2 & 65.2 & 69.1 & 69.3 & 82.2 & 80.0 & - & -\\
& \it Headlight & 0.8 & 1.8 & 2.1 & 2.6 & 2.9 & 3.5 & 3.8 & 4.0 & 25.5 & 21.2 & - & -\\ 
& \it Wheel & 15.1 & 19.8 & 18.2 & 23.1 & 22.9 & 27.1 & 27.0 & 27.5 & 50.6 & 47.3 & 5.3 & 4.9\\ 
\hline
\multirow{3}{*}{\it Car} 
& \it Backside & 13.8 & 14.7 & 20.0 & 18.6 & 28.4 & 23.2 & 28.5 & 28.4 & 43.7 & 42.2 & - & -\\
& \it Licence Plate & 15.3 & 15.3 & 12.6 & 15.5 & 15.0 & 20.5 & 20.2 & 21.5 & 41.0 & 38.4 & 5.2 & -\\ 
& \it Wheel & 20.2 & 22.2 & 19.2 & 26.5 & 26.5 & 30.4 & 30.4 & 31.0 & 59.3 & 59.2 & 5.5 & 16.4\\ 
\hline
\multirow{3}{*}{\it Cat} 
& \it Head & 25.4 & 36.9 & 36.6 & 48.8 & 48.2 & 54.7 & 54.1 & 54.6 & 82.2 & 80.5 & 10.9 & -\\
& \it Eye & 10.0 & 10.0 & 10.5 & 16.7 & 16.7 & 21.2 & 21.2 & 21.4 & 45.6 & 43.9 & 1.4 & -\\ 
& \it Tail & 0.1 & 0.4 & 1.1 & 1.5 & 1.6 & 2.3 & 2.2 & 2.4 & 15.5 & 9.8 & - & -\\ 
\hline
\multirow{3}{*}{\it Cow} 
& \it Head & 29.2 & 31.8 & 31.8 & 39.2 & 39.3 & 47.1 & 47.3 & 47.9 & 64.9 & 64.9 & 35.3 & -\\
& \it Horn & 3.9 & 6.1 & 7.2 & 10.2 & 10.8 & 14.5 & 14.9 & 15.1 & 23.1 & 22.0 & - & -\\ 
& \it Muzzle & 7.8 & 9.9 & 12.1 & 17.1 & 17.4 & 20.4 & 20.8 & 21.1 & 41.6 & 40.7 & - & -\\ 
\hline
\multirow{3}{*}{\it Dog} 
& \it Head & 30.1 & 32.7 & 32.8 & 42.3 & 42.9 & 56.1 & 56.5 & 56.4 & 79.1 & 79.9 & 20.9 & -\\
& \it Eye & 6.5 & 6.5 & 7.1 & 8.1 & 8.9 & 10.6 & 10.6 & 10.9 & 17.8 & 16.4 & - & -\\ 
& \it Tail & 0.1 & 0.2 & 0.8 & 1.2 & 1.1 & 1.4 & 1.5 & 1.7 & 9.4 & 6.4 & - & -\\ 
\hline
\multirow{3}{*}{\it Horse} 
& \it Head & 30.7 & 33.8 & 33.5 & 35.7 & 34.9 & 37.9 & 37.2 & 37.4 & 64.4 & 63.9 & 22.2 & -\\
& \it Torso & 8.1 & 16.0 & 46.1 & 47.2 & 53.2 & 48.4 & 55.7 & 59.4 & 71.9 & 68.9 & - & -\\ 
& \it Leg & 0.6 & 12.0 & 14.7 & 4.6 & 5.3 & 4.6 & 6.9 & 7.1 & 14.3 & 11.5 & - & -\\ 
\hline
\multirow{3}{*}{\it Sheep} 
& \it Head & 25.4 & 28.2 & 28.6 & 33.1 & 33.5 & 35.2 & 35.7 & 35.6 & 49.2 & 49.0 & 18.4 & -\\
& \it Horn & 1.9 & 2.9 & 3.3 & 2.7 & 3.2 & 3.1 & 3.5 & 3.6 & 27.3& 23.1 & - & -\\ 
& \it Leg & 0.4 & 1.1 & 2.4 & 3.1 & 3.5 & 3.2 & 3.6 & 3.8 & 13.3 & 11.6 & - & -\\ 
\hline
\multicolumn{2}{|c|}{\bf {\it  mAP}} & \bf 12.9 & \bf 15.7 & \bf 18.3 & \bf 22.0 & \bf 23.5 & \bf 25.4 & \bf 26.8 & \bf 27.2 &\bf 43.9 & \bf 41.5 & - & -\\
\hline
\end{tabular}}
}
\end{table*} 

\vspace{-2mm}
\subsection{Part localization} 
\label{sec:part_loc}
\vspace{-1mm}
In this section we evaluate how good our part models are at localizing parts in novel images.
We evaluate part localization in terms of average precision (AP) on the PASCAL-Part \texttt{\small validation} set (which was never seen by our learning procedure). As in \cite{chen14cvpr}, a part is considered correctly localized if is has an $IoU \geq 0.4$ with a ground-truth bounding-box. 

Our location models are conditioned on the object viewpoint, which is however unknown for the objects in \texttt{\small validation}.
We apply our viewpoint classifier $\mathcal{V}^2$ on all objects in \texttt{\small validation} to select what location model to use. When detecting parts we use a linear combination of the score given by the appearance and location models ($\mathcal{A}+\mathcal{L})$.  

We evaluate each component of our system at each stage of the learning, from $\mathcal{T}^0$ to $\mathcal{T}^3$. 
For comparison, we trained additional part appearance models $\mathcal{A}^0$ directly on images retrieved by Google, \emph{before} fitting a bounding-box around each training instance (sec.~\ref{sec:data}). 
\begin{figure*}[t]
  \centering
    \includegraphics[width=\textwidth]{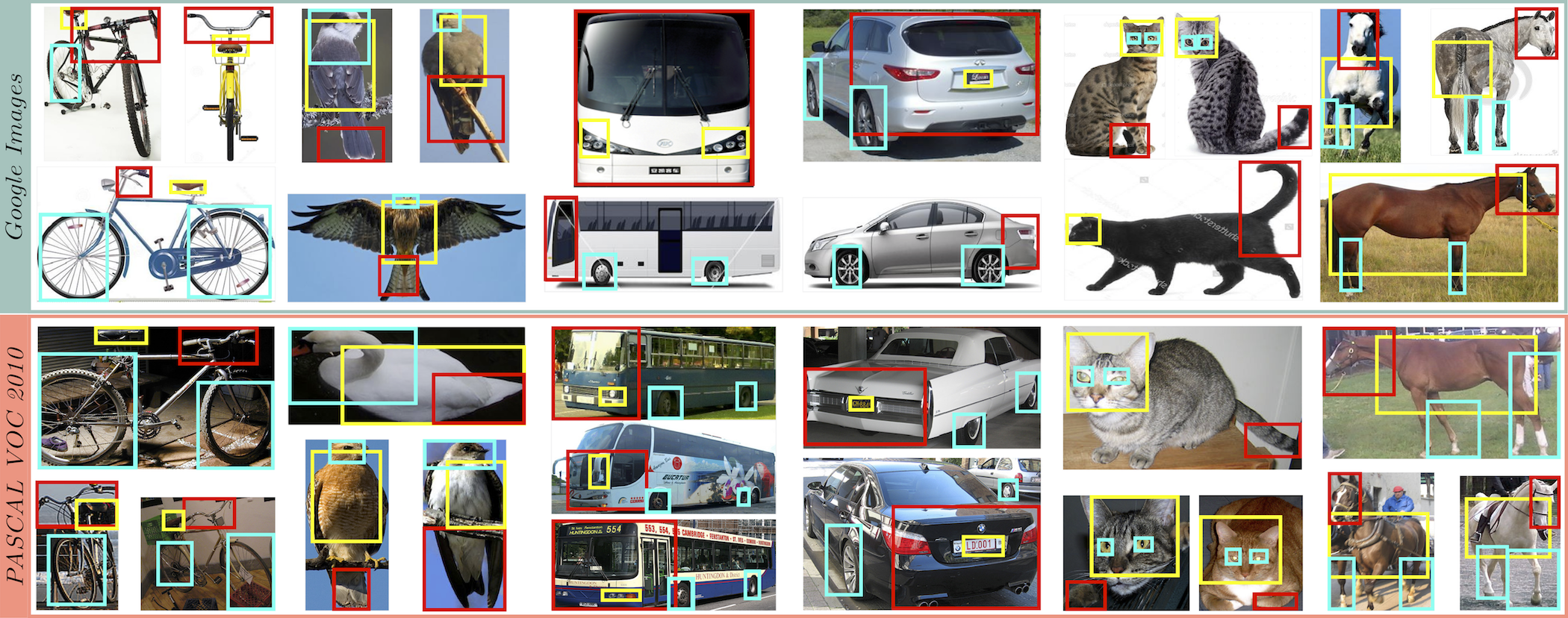}
      \caption{\it \small Detections obtained by running $\mathcal{A}^1+\mathcal{L}^2$ on object images from Google (top) and $\mathcal{A}^3+\mathcal{L}^3$ on objects from PASCAL-Parts (bottom).\vspace{-4mm}} \vspace{-2mm}
      \label{fig:res}
\end{figure*} 
Results are presented in table~\ref{table:res} and fig.~\ref{fig:res}.
Naively using images as returned by Google Images ($\mathcal{A}^0$) leads to an AP of only 12.9.
This reveals how challenging is the task of localizing object parts on a dataset like PASCAL VOC. 
Our refined models $\mathcal{A}^1$ perform already better and improve $\mathcal{A}^0$ by +2.8, showing that our polishing process is useful and provides cleaner examples that lead to better performance. 
The really interesting leap however is achieved in stage $\mathcal{T}^2$ when our framework associates the object to its parts and learns the connection. More precisely, learning the location of the parts under the different viewpoints increases AP to 18.3 ($\mathcal{A}^1 +\mathcal{L}^2$).
Using this information to mine for more part instances and update the appearance model improves performance even further to 22.0 ($\mathcal{A}^2$). Ultimately, the combination of these two models ($\mathcal{A}^2 +\mathcal{L}^2$) brings the performance to 23.5.
This is almost double the initial AP of naively training detectors directly from part images ($\mathcal{A}^0$). 
Importantly, at this point we have a complete class model (appearance, location, viewpoint) trained entirely and automatically from Google Images.
Finally, if we additionally migrate to the PASCAL VOC domain ($\mathcal{T}^3$) and adapt appearance and location models to it, the performance further improves to a final AP of 27.2 ($\mathcal{A}^3+\mathcal{L}^3$). 
The steady improvement exhibited from stage $\mathcal{T}^0$ to $\mathcal{T}^3$ by our incremental learning framework demonstrates its potential in learning complex part models automatically. 

For reference, we train two fully supervised models: $\mathcal{A}^{FS1}$ and $\mathcal{A}^{FS2}$. The former uses manual part bounding-boxes from PASCAL-Part \texttt{\small train}, while the latter also uses the part instances from $\mathcal{T}^0$ collected from the web.
Similarly to sec.~\ref{sec:app}, for each class we took AlexNet CNN and replaced its last layer with a $(P+1)$-way $\mathtt{fc8}$ ($P$ parts an one background class).
These models provide an upper-bound on what can be achieved by any weakly supervised procedure on this dataset.
Our final part detector achieves 27.2 AP, which is 62\% of the performance of $\mathcal{A}^{FS1}$. This is very encouraging, given that we train without part location annotations, whereas $\mathcal{A}^{FS1}$ trains from 15K part bounding-boxes on PASCAL-Part \texttt{\small{train}} (covering our 9 classes with 3 parts each). These take a lot of time as the parts are small and difficult to annotate.
Interestingly, $\mathcal{A}^{FS2}$ performs a little worse than $\mathcal{A}^{FS1}$, despite being trained from more data. We attribute this to the difference between the type of images in PASCAL-Part and on the web.

\noindent{\bf Comparison to LEVAN~\cite{divvala14lcvpr} and NEIL~\cite{chen13iccv}.}
LEVAN and NEIL learn detectors from the web and their original papers do not present quantitative evaluation on part detection. Nonetheless, a few of their models represent semantic parts. We evaluate them in this section, using their DPM models~\cite{felzenszwalb10pami} they released online~\cite{divvala14lcvpr,chen13iccv}.
LEVAN learns multi-component object class detectors.
The components within each object model are labelled with a name, like 'horse jumping' or 'horse head'.
We downloaded the detectors for our nine object classes and selected all components matching our parts. For example, to detect \emph{car licence plate} we run the models labelled as 'plate\_car\_super3' and 'plate\_car\_super6'.
NEIL, instead, learns a collection of separate models, some representing object classes and others part classes, as well as part-of relation between them.
We downloaded all NEIL's models and selected those in a part-of relation with any of the object classes we consider.
This only matches one part \emph{wheel}. Only one generic wheel model is available, not associated to a specific object class.

We run all these part models on PASCAL-Part \texttt{\small validation} and show results in table~\ref{table:res} (rightmost two columns). Note how most of the parts we consider are missing from the components learned by NEIL and LEVAN.
On the few parts that they learned, our part detectors outperform LEVAN and NEIL by a large margin.
The main reason is that their models are trained from part instances downloaded from the web with no (or minimal) refinement: LEVAN uses instances similar to our $\mathcal{T}^0$, and NEIL uses something in between our $\mathcal{T}^1$ and $\mathcal{T}^2$.
A second reason is that LEVAN and NEIL's components are based on simple HOG features, which are weaker than CNNs.

\vspace{-4mm}
\subsection{Object detection} 
\label{sec:det}
\vspace{-1mm}
In this section we augment the R-CNN object class detector~\cite{girshick14cvpr} with our part models. The standard R-CNN detector scores each object proposal $w$ in an image with a root filter $\mathcal{R}$ covering the whole object.
Inspired by~\cite{felzenszwalb10pami} we add a collection of parts arranged in a deformable configuration:
\begin{equation}
\small
\vspace{-2mm}
\text{score}(w) = \mathcal{R}(w) + \sum_{i}^{N} \max_{w' \in \Upsilon} (\alpha_{i} \cdot \mathcal{A}_i(w') + \beta_{i} \cdot \mathcal{L}_{i, \mathcal{V}(w)}(w'))
\label{eq:obj_det}
\end{equation}
where $\Upsilon$ is the set of part proposals inside $w$. For each part $i$, the $\max$ operation looks for the best fitting proposal $w'\in \Upsilon$ according to the part appearance model ($\mathcal{A}_i$) and location model ($\mathcal{L}_i$), 
measuring how likely part $i$ is to appear at the location $w'$.
As we have a separate location model per viewpoint, we use our classifier $\mathcal{V}$ to select which one to use on $w$.
We use the same set of proposals for both objects and parts (sec.~\ref{sec:data}).
We set the weights $\alpha$ and $\beta$ by cross-validation on \texttt{\small train}.
%
This overall object class model is similar to~\cite{felzenszwalb10pami}, but instead of using Gaussian part location models, we have a full probability distributions given by kernel density estimators (eq.~\ref{eq:lm}).

\begin{table}
\caption{\small Object detection results (average precision). \vspace{-2mm}} \vspace{-1mm}  
\label{table:objLoc}
\footnotesize
\resizebox{\columnwidth}{!}{
\begin{tabular}{l  : c : c c c c c c c c c : c}
\toprule
\multirow{2}{*}{\it Model} & Test & \multirow{2}{*}{\it Bicycle} & \multirow{2}{*}{\it Bird} & \multirow{2}{*}{\it Bus} & \multirow{2}{*}{\it Car} & \multirow{2}{*}{\it Cat} & \multirow{2}{*}{\it Cow} & \multirow{2}{*}{\it Dog} & \multirow{2}{*}{\it Horse} & \multirow{2}{*}{\it Sheep} & \multirow{2}{*}{mean} \\
& VOC & & & & & & & & & & \\\midrule
R-CNN$_1$ & 2010 & 64.6 & 46.8 & 63.5 & 56.3 & 69.0 & 45.4 & 62.4 & 55.1 & 54.8 & 57.4 \\
R-CNN$_2$ & 2010 & 64.1 & 43.7 & 62.5 & 56.1 & 67.3 & 45.1 & 61.2 & 55.0 & 54.2 & 56.6 \\
R-CNN$_1$ + parts & 2010 & 66.9 & 49.3 & 65.6 & 58.4 & 70.8 & 46.8 & 64.3 & 58.0 & 55.9 & \bf 59.6\\\hline
R-CNN$_1$ & 2012 & 63.5 & 44.4 & 62.2 & 55.5 & 68.1 & 44.6 & 61.0 & 53.5 & 55.4 & 56.5 \\
R-CNN$_1$ + parts & 2012 & 66.1 & 47.2 & 64.1 & 58.0 & 69.6 & 45.9 & 62.8 & 56.7 & 56.4 & \bf 58.5\\
\bottomrule
\end{tabular}}
\vspace{-3mm}
\end{table}

We train the root filer on PASCAL-Parts \texttt{\small train} as in~\cite{girshick14cvpr}.
The other elements of the model are learned from the web and PASCAL-Parts using our technique (sec.~\ref{sec:tech}),
i.e. $\mathcal{A}^3$ as part filters, $\mathcal{L}^3$ as location models and $\mathcal{V}^2$ as viewpoint classifiers.
No manual part location annotations is used for training.
We report object detection results on \texttt{\small validation} of PASCAL VOC 2010 and 2012, in terms of AP (table~\ref{table:objLoc}).
Compared to using the R-CNN root filter alone (R-CNN$_1$), adding parts increases its performance by 2-3\% on all classes, and on both test sets (R-CNN$_1+$ parts). This shows that our part models can help object class detection, even when added to an already strong fully supervised detector like R-CNN. 
This is an interesting result, especially considering that our parts are designed to be {\em semantic},
as opposed to discriminative arbitrary patches~\cite{endres13cvpr,felzenszwalb10pami}.

For a fully fair comparison, we also train another R-CNN model on object instances from both PASCAL-Part {\em and} the images we downloaded from the web ($\mathcal{T}^0$). Interestingly, training using this additional data decreases performance by 0.8\% (R-CNN$_2$). Again, we attribute this to the difference between the type of images in PASCAL-Part and on the web.



\vspace{-3mm}
\section{Conclusions}
\label{sec:conl}
\vspace{-1mm}
We presented a technique for learning part-based models from the web. It operates by collecting object and part instances and by automatically connecting them in an incremental learning procedure. Our models encompass the appearance of parts and their spatial arrangement on the object, specific to each viewpoint.
We reported results on the challenging PASCAL-Parts which show that our technique is able to learn good part detectors from the web. Finally, we demonstrated the value of our part models by enriching the R-CNN object detector with parts.


\noindent{\bf Acknowledgments.} Support by ERC Starting Grant VisCul.
\ifCLASSOPTIONcaptionsoff
  \newpage
\fi



%
\vspace{-3mm}
\bibliographystyle{IEEEtran}
\bibliography{../../bibtex/shortstrings,../../bibtex/vggroup,../../bibtex/calvin}




%








\end{document}